%% file: main.tex
\documentclass[letterpaper]{article} 
\usepackage[preprint]{aaai2027}  
\usepackage[hyphens]{url}  
\usepackage{graphicx} 
\usepackage{natbib}  
\usepackage{caption} 
\usepackage{algorithm}
\usepackage{algorithmic}

\usepackage{newfloat}
\usepackage{listings}
\DeclareCaptionStyle{ruled}{labelfont=normalfont,labelsep=colon,strut=off} 
\floatstyle{ruled}
\newfloat{listing}{tb}{lst}{}
\floatname{listing}{Listing}

\usepackage{booktabs}

\usepackage{caption}
\usepackage{amsmath}
\usepackage{algorithm, algorithmic}
\usepackage{amssymb}
\usepackage[table]{xcolor}
\usepackage{multirow}
\usepackage{xcolor}
\definecolor{mygray}{gray}{0.66}
\newcommand{\mygray}[1]{\textcolor{mygray}{#1}}

\newcommand{\myurl}[1]{\textcolor{blue}{\ttfamily{\url{#1}}}}
\title{
Learning How the World Evolves:\\
Extrapolative Video World Models via Latent Dynamics Reasoning
}
\author{
{
Haodong Li\textsuperscript{\rm 1,2}
\ Shaoteng Liu\textsuperscript{\rm 2}\thanks{Project lead.}
\ Tianyu Wang\textsuperscript{\rm 2}
\ Chongjian Ge\textsuperscript{\rm 2}
\ Sihui Ji\textsuperscript{\rm 2}
\ Jiahan Zhang\textsuperscript{\rm 2}
\ Xin Lin\textsuperscript{\rm 1,2}
\ Haolin Lu\textsuperscript{\rm 1}
\ Zhe Lin\textsuperscript{\rm 2}
\ Manmohan Chandraker\textsuperscript{\rm 1}\\
\textsuperscript{\rm 1}UCSD
\ \textsuperscript{\rm 2}Adobe
}
}
\affiliations{

}

\begin{document}

\input{Figs/Teaser}
\maketitle

\input{Secs/Abs}
\input{Secs/Intro}

\input{Figs/Pipeline}
\input{Secs/Related_works}

\input{Secs/Method}

\input{Secs/Experiments}

\input{Secs/Summary}

\bibliography{aaai2027}


\end{document}

%% file: Secs/Abs.tex
\begin{abstract}

The world evolves following its dynamics, i.e., its laws of motion.
However, leading video diffusion models largely fit the pixels without modeling how the pixels transit over time.
Thus, they render visually plausible frames but may not accurately obey the laws.
To capture the dynamics purely from pixels, we introduce Latent Dynamics Reasoning (LDR).
LDR casts the latent transition as an explicit kinematic integration, where the lower-order dynamics are integrated numerically and the model regresses only the third- and higher-order residual that drives the rollout.
For this integration to extrapolate better, LDR runs it on a structured latent rather than dense convolutional features.
Following PhyWorld~\cite{kang2024far}, we validate LDR on a controlled white-box physics benchmark spanning five tasks (uniform motion, parabola, collision, bouncing, looming), focusing on out-of-distribution scenarios that reveal whether a model has truly learned the underlying dynamics.
LDR extrapolates the learned dynamics far better: the gap between its in- and out-of-distribution error is over 20$\times$ smaller than the video diffusion baseline's, under both single- and joint-task training at 256$^2$ resolution, while using 26$\times$ fewer parameters and running 143$\times$ faster.
LDR can even generalize under severe shift: for example, trained only on red balls moving left-to-right, it correctly predicts the motion of a blue square moving right-to-left.
To our knowledge, this is the first video world model that extrapolates learned dynamics beyond its training distribution.
Project page: \myurl{https://lat-dyn-reason.github.io/}.

\end{abstract}

%% file: Secs/Intro.tex
\section{Introduction}
\label{sec:intro}

The world evolves following its dynamics, i.e., the laws of motion that govern how its state changes over time.
However, leading video diffusion models mainly learn ``what the world looks like'', without capturing ``how it evolves'', i.e., the underlying dynamics that drive the transitions of pixels.
Thus, they render visually plausible frames but may not accurately obey the laws.
We argue that capturing the underlying dynamics from pixels is one of the most fundamental differences that distinguish video world models from video generators.
A video world model should capture how the world evolves and accurately extrapolate the learned dynamics to unseen scenarios.

We introduce Latent Dynamics Reasoning (LDR), which predicts future frames by reasoning about the latent dynamics rather than regressing them directly (Fig.~\ref{fig:teaser}A).
Specifically, LDR casts the latent transition as an explicit kinematic integration (Fig.~\ref{fig:pipeline}B).
From the structured latent (SL) of each conditioning frame, LDR starts by forming the first two time derivatives to initialize the rollout.
It then rolls out step by step: the model regresses only the third- and higher-order residual, then numerically integrates the second-, first-, and zero-order SL in turn.
This forces the model to learn the underlying dynamics, i.e., how the latent evolves over time, rather than merely what the next latent is.
The SL gives a compact, structured representation free of the redundant semantic and appearance information carried by dense convolutional features, which makes the differentiation and integration of the dynamics more stable and more reliable when extrapolating (Fig.~\ref{fig:pipeline}A).
Finally, LDR decodes each future frame from its SL by warping the conditioning frame (Fig.~\ref{fig:pipeline}C).

We build our benchmark using the simulator of PhyWorld~\cite{kang2024far}, a clean and controlled testbed, and validate LDR on five tasks: uniform motion, parabola, collision, bouncing, and looming.
For each task, we define in-distribution (ID) ranges of the initial conditions and out-of-distribution (OOD) ranges that share the same laws of motion.
The model is trained only on ID samples.
In ID testing, the model only needs to reproduce motions it has seen.
But in OOD testing, the model is required to extrapolate the learned dynamics beyond the training distribution, which cleanly distinguishes capturing the dynamics from merely memorizing the pixels.
In addition, because the simulator is white-box, we can directly measure the accuracy of the learned dynamics by parsing each object's position and size from the predicted pixels and comparing them against the ground truth (GT).

Compared with PhyWorld's standard video diffusion baseline (a DiT, for which we choose DiT-S), LDR extrapolates the learned dynamics to OOD samples far better while being much smaller and faster.
Averaged over the five tasks at 256$^2$ resolution, LDR's gap between ID and OOD error is over 20$\times$ smaller than the baseline's under both single-task and joint five-task training, while using 26$\times$ fewer parameters and running 143$\times$ faster.
This efficiency comes from predicting future frames in a single forward pass, without iterative sampling or test-time optimization.
We further ablate LDR's two components by removing the dynamics reasoning or replacing SL with dense convolutional features.
Either ablation widens the ID-OOD gap several-fold, confirming that both are necessary.
Qualitatively, LDR tracks the true motion under both single-task and joint training where the DiT baseline and ablated variants drift (Fig.~\ref{fig:qualitative-s},~\ref{fig:qualitative-j}).
LDR can also generalize the learned dynamics under large OOD shifts (Fig.~\ref{fig:teaser},~\ref{fig:severe_shift}).
Our contributions are summarized as follows:
\begin{itemize}
\item To learn the dynamics behind the pixels and extrapolate them to unseen scenarios, we propose Latent Dynamics Reasoning (LDR). To our knowledge, this is the first video world model that extrapolates learned dynamics beyond its training distribution.\footnote{
Scope: We validate LDR as a principle on simulated scenarios with simple objects. Scaling to richer, even real-world scenes with larger models is future work.
}
\item We instantiate LDR as a model that reasons about latent dynamics through kinematic integration in structured latent space.
On a controlled benchmark of five physics tasks, LDR extrapolates far better than the video diffusion baseline while being much smaller and faster.
\end{itemize}

%% file: Figs/Pipeline.tex
\begin{figure*}[!ht]
    \centering
    \includegraphics[width=\linewidth]{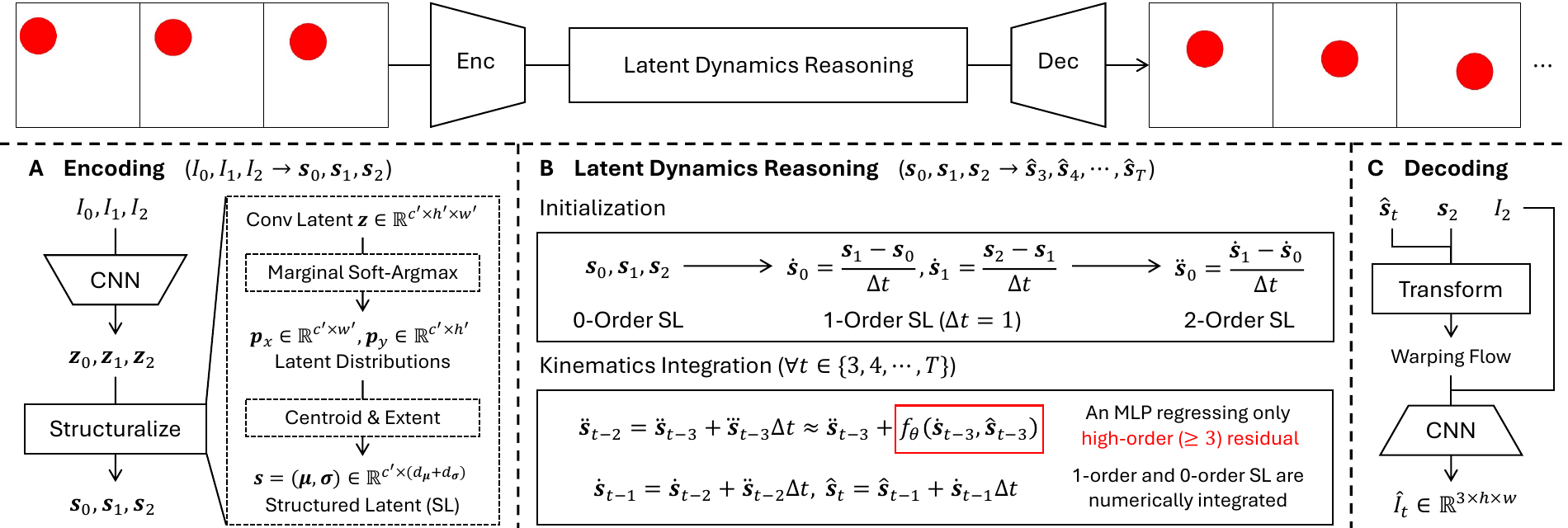}
    \vspace{-.6cm}
\caption{
\textbf{Overview of LDR.}
Given three conditioning frames $I_0,I_1,I_2$, LDR predicts the future frames $\hat{I}_3,\hat{I}_4,\dots,\hat{I}_T$ in three stages.
\textbf{(A)} LDR first encodes each input frame into a structured latent (SL).
\textbf{(B)} LDR measures the low-order time derivatives of the given SL, then rolls out by regressing only the high-order residual with $f_\theta$ and numerically integrating the lower orders to the next latent $\hat{\boldsymbol{s}}_t$ ($t\in\{3,4,\cdots,T\}$).
\textbf{(C)} LDR finally decodes each predicted latent to an RGB frame $\hat{I}_t$.
}
    \label{fig:pipeline}
    \vspace{-.2cm}
\end{figure*}

%% file: Secs/Related_works.tex
\section{Related Works}
\label{sec:rw}

\subsubsection{Video Generation and Video World Models.}
Video generation models synthesize realistic frames via latent diffusion modeling \cite{ho2020ddpm, song2021ddim, song2021scoresde, rombach2022ldm, peebles2023dit, blattmann2023svd, brooks2024sora, yang2024cogvideox, li2026rolling}.
Some of them are regarded as video world models \cite{ha2018worldmodels, hafner2023dreamerv3, bruce2024genie, alonso2024diamond, agarwal2025cosmos, agarwal2026cosmos}.
Realism, though, does not imply that the underlying dynamics is captured: the latent transition is a black box fit to the training distribution, and Cosmos \cite{agarwal2025cosmos, agarwal2026cosmos} also reports that physical accuracy remains unsolved.
Both JEPA \cite{bardes2024vjepa, assran2025vjepa2} and our LDR can be categorized as ``next-latent prediction'', but like DiT, JEPA regresses the future latents directly without reasoning about the dynamics.
We argue that capturing and extrapolating the learned dynamics is what should distinguish video world models from video generation models, and that is the capability LDR provides.

\subsubsection{Learning and Extrapolating the Laws of Motion.}
Learning the laws of motion has been studied extensively.
One route captures dynamics accurately but relies on external signals: some integrate dynamics over known states \cite{battaglia2016interaction, sanchez2020learning, liu2024segno, kipf2018neural, lam2023learning, yin2023continuous, ruhling2023dyffusion}, some under conservation or PDE priors \cite{greydanus2019hnn, cranmer2020lnn, guen2020phydnet, alet2021noether}, while others draw supervision from physics engines or physical laws \cite{xue2024phyt2v, lin2025reasoning, li2026phi}.
Another route learns dynamics purely from pixels via neural ODE modeling \cite{chen2018neuralode, rubanova2019latentode, park2021vidode, watter2015e2c, krishnan2015dkf, yildiz2019ode2vae}, yet these are validated only within the training regime, leaving extrapolation of the learned dynamics largely unexplored.

Beyond its training distribution, a neural network extrapolates according to the inductive bias of its architecture rather than the data it has seen \cite{xu2021how}.
A model therefore extrapolates dynamics only when the reasoning of dynamics is built into its architecture rather than fit directly from data.
Following this principle, LDR reasons about the underlying dynamics through kinematic integration in a structured latent space, aiming not only to capture the dynamics purely from pixels but also to extrapolate the learned dynamics beyond the training distribution.
To our knowledge, LDR is the first video world model that extrapolates learned dynamics beyond its training distribution.

\subsubsection{Structured Representation.}
As mentioned, LDR reasons about the dynamics in a structured latent space.
This choice is motivated by \cite{jakab2018unsupervised, kulkarni2019unsupervised, minderer2019unsupervised, daniel2024ddlp, locatello2020object, wu2023slotformer, kipf2020contrastive, jiang2024slot}, where structured representations (e.g., geometric coordinates or object-centric slots) outperform unstructured features (e.g., convolutional features) in sequential modeling.
For more stable reasoning and more reliable extrapolation, LDR follows \cite{jakab2018unsupervised, kulkarni2019unsupervised, minderer2019unsupervised} to represent each frame as the geometric coordinates of its convolutional feature.

%% file: Secs/Method.tex
\section{Method}
\label{sec:method}

Latent Dynamics Reasoning (LDR) predicts future frames by reasoning about the dynamics behind the pixels (Fig.~\ref{fig:pipeline}).
It first encodes each input frame into a structured latent (SL) (Fig.~\ref{fig:pipeline}A).
It then reasons about how this latent evolves via explicit kinematic integration (Fig.~\ref{fig:pipeline}B).
Finally, it decodes each predicted latent back to a frame (Fig.~\ref{fig:pipeline}C).
Although LDR consists of three conceptual stages, it predicts in a single feed-forward pass, without test-time optimization or iterative solvers.

\subsection{Latent Dynamics Reasoning}

From the conditioning latents, LDR first measures their first- and second-order time derivatives to initialize the rollout.
It then regresses only the third- and higher-order residual, and rolls out the future latents by kinematic integration.

\subsubsection{Initialization.}
Because LDR regresses only the third- and higher-order residual, it initializes the rollout only up to the second-order derivative.
A second-order finite difference is fixed by three points, so LDR conditions on three frames.
It forms the initial dynamics from the conditioning latents by finite differences:
\begin{equation}
\dot{\boldsymbol{s}}_0 = \frac{\boldsymbol{s}_1-\boldsymbol{s}_0}{\Delta t},\
\dot{\boldsymbol{s}}_1 = \frac{\boldsymbol{s}_2-\boldsymbol{s}_1}{\Delta t},\
\ddot{\boldsymbol{s}}_0 = \frac{\dot{\boldsymbol{s}}_1-\dot{\boldsymbol{s}}_0}{\Delta t},
\label{eq:init}
\end{equation}
with $\Delta t=1$.
If we consider $\boldsymbol{s}$ itself as ``position'', $\dot{\boldsymbol{s}}$ and $\ddot{\boldsymbol{s}}$ are its ``velocity'' and ``acceleration'', both derivatives of the latent trajectory.

\subsubsection{Kinematics Integration.}
LDR then rolls out the next latent one at a time, regressing the third- and higher-order residual and integrating the lower orders, $\forall t\in\{3,\dots,T\}$:
\begin{equation}
\begin{aligned}
\ddot{\boldsymbol{s}}_{t-2} &= \ddot{\boldsymbol{s}}_{t-3} + \dddot{\boldsymbol{s}}_{t-3}\Delta t
  \approx \ddot{\boldsymbol{s}}_{t-3} + f_\theta(\dot{\boldsymbol{s}}_{t-3}, \hat{\boldsymbol{s}}_{t-3}),\\
\dot{\boldsymbol{s}}_{t-1} &= \dot{\boldsymbol{s}}_{t-2} + \ddot{\boldsymbol{s}}_{t-2}\Delta t,\
\hat{\boldsymbol{s}}_{t} = \hat{\boldsymbol{s}}_{t-1} + \dot{\boldsymbol{s}}_{t-1}\Delta t.
\end{aligned}
\label{eq:rollout}
\end{equation}
The rollout is initiated by the conditioning latents, i.e., $\hat{\boldsymbol{s}}_i=\boldsymbol{s}_i$ for $i\in\{0,1,2\}$.
Here $f_\theta(\cdot)=\tanh(\mathrm{MLP}(\cdot))$ regresses the third- and higher-order residual, i.e., the change in the second-order latent, approximating $\dddot{\boldsymbol{s}}_{t-3}\Delta t$.
The integration then propagates this residual down through the second- and first-order latents to produce the next latent.
Only $f_\theta$ is learned; the integration chain is fixed (Alg.~\ref{alg:ldr}).

\input{Algs/LDR}

\subsubsection{Why LDR Extrapolates?}
When tested beyond the training distribution, a network follows the inductive bias of its architecture rather than the data it was trained on, and a plain regressor with no such bias flattens toward the training mean off-support \cite{xu2021how}.
In OOD video prediction, this manifests as reproducing the closest training example instead of accurately obeying the underlying dynamics \cite{kang2024far}.
LDR instead casts the state transition as an explicit kinematic integration, forcing the model to learn the abstract pattern of the state's motion, i.e., ``how the state evolves'' rather than merely ``what the next state is''.
This architecturally builds an inductive bias toward capturing the underlying dynamics behind the pixels, letting LDR extrapolate them well beyond the training distribution.

\subsection{Structured Latent}

The more the latent state entangles with dynamics-irrelevant details, the harder the dynamics is to reason (i.e., differentiation and integration) and to extrapolate.
LDR thus reasons in the SL space, a compact representation free of the dynamics-irrelevant detail (e.g., appearance and semantics) carried by dense convolutional features, which makes the dynamics reasoning more stable and more reliable when extrapolating.

\subsubsection{Encoding.}
Before dynamics reasoning, LDR first encodes each input frame into a structured latent, $\forall i\in\{0,1,2\}$:
\begin{equation}
\boldsymbol{s}_i = E_\phi(I_i)= \operatorname{Struct}\left(\mathcal{C}_\phi\left(I_i\right)\right) = (\boldsymbol{\mu}_i, \boldsymbol{\sigma}_i).
\label{eq:encode}
\end{equation}
A convolutional network $\mathcal{C}_\phi$ maps the frame $I_i$ to a feature map $\boldsymbol{z}_i$.
A marginal soft-argmax then turns each channel of $\boldsymbol{z}_i$ into a spatial distribution, from which LDR extracts the geometric coordinate (i.e., the centroid $\boldsymbol{\mu}_i$ and extent $\boldsymbol{\sigma}_i$) and forms the structured latent $\boldsymbol{s}_i$ \cite{jakab2018unsupervised,kulkarni2019unsupervised,minderer2019unsupervised}.

\subsubsection{Decoding.}
After dynamics reasoning, LDR decodes each predicted latent back to an RGB frame by warping the conditioning frame, following \cite{siarohin2019fomm,siarohin2021mraa,gao2019disentangling}, $\forall t\in\{3,\dots,T\}$:
\begin{equation}
\hat I_t = D_\psi(\hat{\boldsymbol{s}}_t, \boldsymbol{s}_2, I_2) = \mathcal{R}_\psi(I_2, \mathcal{T}_\psi(\hat{\boldsymbol{s}}_t, \boldsymbol{s}_2)).
\label{eq:decode}
\end{equation}
A learned transformation module $\mathcal{T}_\psi$ predicts a dense warping flow from the conditioning latent $\boldsymbol{s}_2$ to the predicted one $\hat{\boldsymbol{s}}_t$.
$\mathcal{R}_\psi$ then predicts the conditioning frame $I_2$ along this flow and renders frame $\hat I_t$.

\subsection{Optimization}

We train all components of LDR (the encoder $E_\phi$, the decoder $D_\psi$, and the high-order dynamics residual predictor $f_\theta$) jointly and from scratch: the model carries no pretrained weights or modules.
Three terms supervise the training.
An RGB reconstruction term optimizes the encoder and decoder:
$
\mathcal{L}^{\text{rgb}}_{\text{ae}} = \sum\nolimits_{t=0}^{T}\lVert \Phi(D_\psi(E_\phi(I_t), \boldsymbol{s}_2, I_2)) - \Phi(I_t)\rVert_1
$,
where $\Phi$ is a multi-scale image feature extractor \cite{johnson2016perceptual}.
An RGB rollout term optimizes the entire model:
$
\mathcal{L}^{\text{rgb}}_{\text{roll}} = \sum\nolimits_{t=3}^{T}\lVert \Phi(\hat I_t) - \Phi(I_t)\rVert_1
$.
A latent rollout term primarily optimizes $f_\theta$: 
$
\mathcal{L}^{\text{SL}}_{\text{roll}} = \sum\nolimits_{t=3}^{T}\lVert \hat{\boldsymbol{s}}_t - \operatorname{sg}(\boldsymbol{s}_t)\rVert_2^2
$.
The full objective is:
\begin{equation}
\mathcal{L}_{\text{LDR}} = \mathcal{L}^{\text{rgb}}_{\text{roll}} + \lambda^{\text{rgb}}_{\text{ae}}\mathcal{L}^{\text{rgb}}_{\text{ae}} + \lambda^{\text{SL}}_{\text{roll}}\mathcal{L}^{\text{SL}}_{\text{roll}}.
\label{eq:loss}
\end{equation}
In training, we grow the rollout horizon from short to full, which stabilizes long-horizon backpropagation.
In testing, we always roll out the full horizon.

%% file: Algs/LDR.tex
\begin{algorithm}[t]
\caption{Latent Dynamics Reasoning}
\label{alg:ldr}
\begin{algorithmic}[1]
\REQUIRE conditioning frames $I_0,I_1,I_2$; horizon $T$; encoder $E_\phi$; decoder $D_\psi$; high-order ($\geq$ 3) dynamics residual regressor $f_\theta$; $\Delta t=1$
\FOR{$i=0$ {\bfseries to} $2$}
\STATE $\boldsymbol{s}_i \gets E_\phi(I_i)$ \hfill// encode
\ENDFOR
\STATE $\displaystyle \dot{\boldsymbol{s}}_0 \gets \frac{\boldsymbol{s}_1-\boldsymbol{s}_0}{\Delta t}$;\ $\displaystyle \dot{\boldsymbol{s}}_1 \gets \frac{\boldsymbol{s}_2-\boldsymbol{s}_1}{\Delta t}$ \hfill// first order
\STATE $\displaystyle \ddot{\boldsymbol{s}}_0 \gets \frac{\dot{\boldsymbol{s}}_1-\dot{\boldsymbol{s}}_0}{\Delta t}$ \hfill// second order
\FOR{$t=3$ {\bfseries to} $T$}
  \STATE $\ddot{\boldsymbol{s}}_{t-2} \gets \ddot{\boldsymbol{s}}_{t-3} + f_\theta(\dot{\boldsymbol{s}}_{t-3}, \hat{\boldsymbol{s}}_{t-3})$ \\// regress high-order residual, get next 2-order latent
  \STATE $\dot{\boldsymbol{s}}_{t-1} \gets \dot{\boldsymbol{s}}_{t-2} + \ddot{\boldsymbol{s}}_{t-2}\Delta t$
  \STATE $\hat{\boldsymbol{s}}_{t} \gets \hat{\boldsymbol{s}}_{t-1} + \dot{\boldsymbol{s}}_{t-1}\Delta t$ \\// integrate to next latent (zero-order)
  \STATE $\hat I_t \gets D_\psi(\hat{\boldsymbol{s}}_{t}, \boldsymbol{s}_2, I_2)$ \hfill// decode
\ENDFOR
\ENSURE predicted frames $\hat I_3,\dots,\hat I_T$
\end{algorithmic}
\end{algorithm}

%% file: Secs/Experiments.tex
\input{Figs/Qualitative-S}
\input{Tabs/main_256_single}
\input{Tabs/main_256_joint}
\input{Tabs/main_128_single}
\input{Tabs/main_128_joint}

\section{Experiments}
\label{sec:exp}

\subsection{Experimental Setup}

\subsubsection{Benchmark.}
For validating LDR, we build a controlled benchmark on the PhyWorld simulator~\cite{kang2024far}, spanning five physics tasks involving one or two moving balls:
\emph{uniform motion} (a ball translating at a constant velocity);
\emph{parabola} (a projectile under gravity);
\emph{collision} (two balls colliding elastically head-on);
\emph{bouncing} (a projectile rebounding off the ground, losing energy at each bounce);
\emph{looming} (a ball translating while growing or shrinking).
In every task, the balls share the same ID range of initial conditions: a speed $v\in[1,4]$ and a radius $r\in[0.7,1.4]$ in a world of scale $10$ (looming further fixes the growing or shrinking rate $|\dot r|\in[0,0.03]$).
OOD test samples push these conditions beyond both ends of the ID range: the speed to $v\in[0.05,6]$, the radius to $r\in[0.6,2]$, and the scale rate to $|\dot r|\in[0.05,0.09]$.
Since the same motion laws hold on both splits, OOD testing effectively evaluates the model's capability in extrapolating the learned underlying dynamics in unseen scenarios.

\subsubsection{Training protocols.}
We train all methods (the DiT-S baseline, LDR, and its ablated variants) from scratch.
Single-task training fits one model per task.
Joint training fits one model on all five tasks at once, which is a harder setting.
Every model conditions on three frames, predicts the next $29$ (i.e., $T=31$), and trains only on ID samples.

\subsubsection{Metrics.}
Because the simulator is white-box and each object in our benchmark is a ball, to directly evaluate the dynamics, we extract each object's center and radius from the predicted frames and compare them against the GT, giving position error (\texttt{pos}) and radius error (\texttt{rad}).
We report numbers tested on both ID and OOD splits and the ID-OOD gap: $\max(0,\mathrm{OOD}-\mathrm{ID})$.

\subsubsection{Baselines and ablated variants.}
We compare against the DiT-S baseline of PhyWorld, which represents the standard video diffusion solution of the community, and two ablations each remove one LDR component.
Removing dynamics reasoning replaces the kinematic integration with a direct residual regression of the next latent (i.e., regressing $\boldsymbol{s}_{t+1}-\boldsymbol{s}_{t}$).
Removing the structured latent (SL) runs the same dynamics reasoning but on a dense convolutional latent instead.

\subsubsection{Implementation details.}
We train every model from scratch with AdamW (learning rate $10^{-4}$, weight decay $0.01$, gradient clipping $1.0$) and a global batch size of $256$, for $10$K steps by default and $20$K for collision, bouncing, and joint training.
LDR uses a three-layer MLP of width $256$ for $f_\theta$ and weights the losses by $\lambda^{\text{rgb}}_{\text{ae}}{=}1.0$ and $\lambda^{\text{SL}}_{\text{roll}}{=}0.5$.
The training rollout gradually grows to the full $29$ frames by step $8$K.
All runs use seed $42$ and report a single run on $8{\times}8$ NVIDIA A100-80G GPUs with PyTorch 2.4 and CUDA 12.1.
Each model is trained in both $128^2$ and $256^2$ resolution.

\subsection{Experimental Results}
\subsubsection{Quantitative Comparison.}
We report the full quantitative comparison at both resolutions in Tab.~\ref{tab:main:256:single},~\ref{tab:main:256:joint},~\ref{tab:main:128:single},~\ref{tab:main:128:joint}.
At $256^2$, LDR's averaged ID-OOD gap in position error is over $20\times$ smaller than the DiT-S baseline's, under both single-task ($23.9\times$) and joint five-task ($27.7\times$) training.
The baseline reproduces ID motion accurately, yet its OOD error explodes, rising from an average of $0.086$ to $0.592$ under joint training, while LDR holds OOD close to ID ($0.050$ to $0.068$).
This is how a regressor with no dynamics bias behaves off the training distribution: it reverts toward the closest training sample \cite{xu2021how,kang2024far}, while LDR reliably extrapolates the learned dynamics to unseen scenarios.

Higher resolution strengthens LDR's performance gain.
From $128^2$ to $256^2$, the DiT-S baseline's average OOD position error grows (joint $0.222$ to $0.592$) while LDR's shrinks (joint $0.114$ to $0.068$), and the same directions hold under single-task training (Tab.~\ref{tab:main:256:single},~\ref{tab:main:128:single}).
The two methods scale in opposite directions.
A capacity-driven regressor fits the training distribution more tightly at higher resolution, so it collapses harder OOD.
LDR instead captures more precise dynamics from higher-resolution frames, so its ID and OOD error improve together.
Explicit dynamics reasoning makes OOD error track ID error: LDR captures the underlying dynamics instead of memorizing the training distribution.

\subsubsection{Ablation Study.}
Dynamics reasoning is the decisive component.
Removing dynamics reasoning is the most damaging ablation.
This variant regresses the residual to the next latent ($\boldsymbol{s}_{t+1}-\boldsymbol{s}_t$) directly instead of integrating the dynamics, similar to the DiT-S baseline that regresses the entire future clip ($I_3,I_4,\cdots,I_T$).
Under single-task training its averaged ID-OOD gap in position error is several times LDR's, $0.168$ against LDR's $0.013$ at $256^2$.
Under joint five-task training it fails even within the training range: unable to disambiguate the five regimes, it misapplies one task's dynamics to another, for example a downward pull on the horizontal uniform motion task, so its averaged ID position error ($0.494$) is an order of magnitude above LDR's ($0.050$) while its OOD error stays large ($0.624$).\footnote{This variant fails on uniform motion and looming in both ID and OOD under joint training.
Thus, its near-zero ID-OOD gap on these two tasks mainly reflects ID failure rather than robust OOD extrapolation.}
Without an inductive bias toward the dynamics, a direct regressor cannot capture the motion even within the training range.

The SL is a strong complement.
Because LDR's decoding is tightly coupled to the SL, we cannot cleanly remove only the ``Structuralize'' operation shown in Fig.~\ref{fig:pipeline}A.
Instead, we replace LDR's whole encoding (Fig.~\ref{fig:pipeline}A) and decoding (Fig.~\ref{fig:pipeline}C) with the same frozen VAE as the DiT-S baseline \cite{rombach2022ldm,kang2024far}, while keeping the dynamics reasoning unchanged.
This variant extrapolates position far better than the baseline and stays second best overall (Tab.~\ref{tab:main:256:single},~\ref{tab:main:256:joint}), with an average position gap of $0.133$ against the baseline's $0.506$ at $256^2$ joint training, and $0.090$ against $0.300$ under single-task training.
In summary, removing either component widens the gap, so both are necessary to LDR.

\subsubsection{Qualitative Comparison.}
Across both the single-task training and joint five-task training, LDR follows the true motion while the DiT-S baseline and both ablations drift (Fig.~\ref{fig:qualitative-s},~\ref{fig:qualitative-j} report the qualitative comparison at $256^2$).
On the OOD cases, LDR's error map stays dark while others' are visibly lighter.

\input{Figs/Qualitative-J}

\input{Figs/Qualitative-OSS}
\input{Tabs/efficiency}

\subsection{Efficiency Study}
\vspace{-.05cm}
Though it consists of three conceptual stages (Fig.~\ref{fig:pipeline}), LDR predicts in a single feed-forward pass, using $26\times$ fewer parameters than the DiT-S baseline and running up to $143\times$ faster at $256^2$ (Tab.~\ref{tab:efficiency}).
It carries only $4.1$M parameters against the baseline's $106.1$M, with no iterative diffusion sampling and no test-time optimization.
The DiT-S baseline instead runs $50$ denoising steps following \cite{kang2024far}, each a full transformer forward whose cost grows quadratically with token count, plus a heavy VAE encoding and decoding.
LDR's extrapolation is thus not simply a matter of scale or compute: it is at once more extrapolative and one to two orders of magnitude smaller and faster.
This is consistent with \cite{kang2024far}: scaling neither the model nor the data helps a model extrapolate the underlying dynamics.

\subsection{Stress Test}
\vspace{-.05cm}
LDR can also generalize under shifts far larger than the OOD ranges defined in our benchmark.
Although it sees only red balls in training, we test it on unseen earth-textured balls (Fig.~\ref{fig:severe_shift}).
Together with Fig.~\ref{fig:teaser}, these illustrations show that LDR still predicts motion that accurately obeys the dynamics even under large OOD shifts, where the DiT-S baseline fails.
This robustness draws on both components of LDR.
The SL encodes each frame as a geometric structure and decodes it by warping the conditioning frame, which grants robustness to unseen appearances.
The dynamics reasoning learns only the high-order dynamics residual, leaving the low-order dynamics measured rather than learned, which grants robustness to unseen initial low-order dynamics.

%% file: Figs/Qualitative-S.tex
\begin{figure}[!t]
    \centering
    \includegraphics[width=0.99\linewidth]{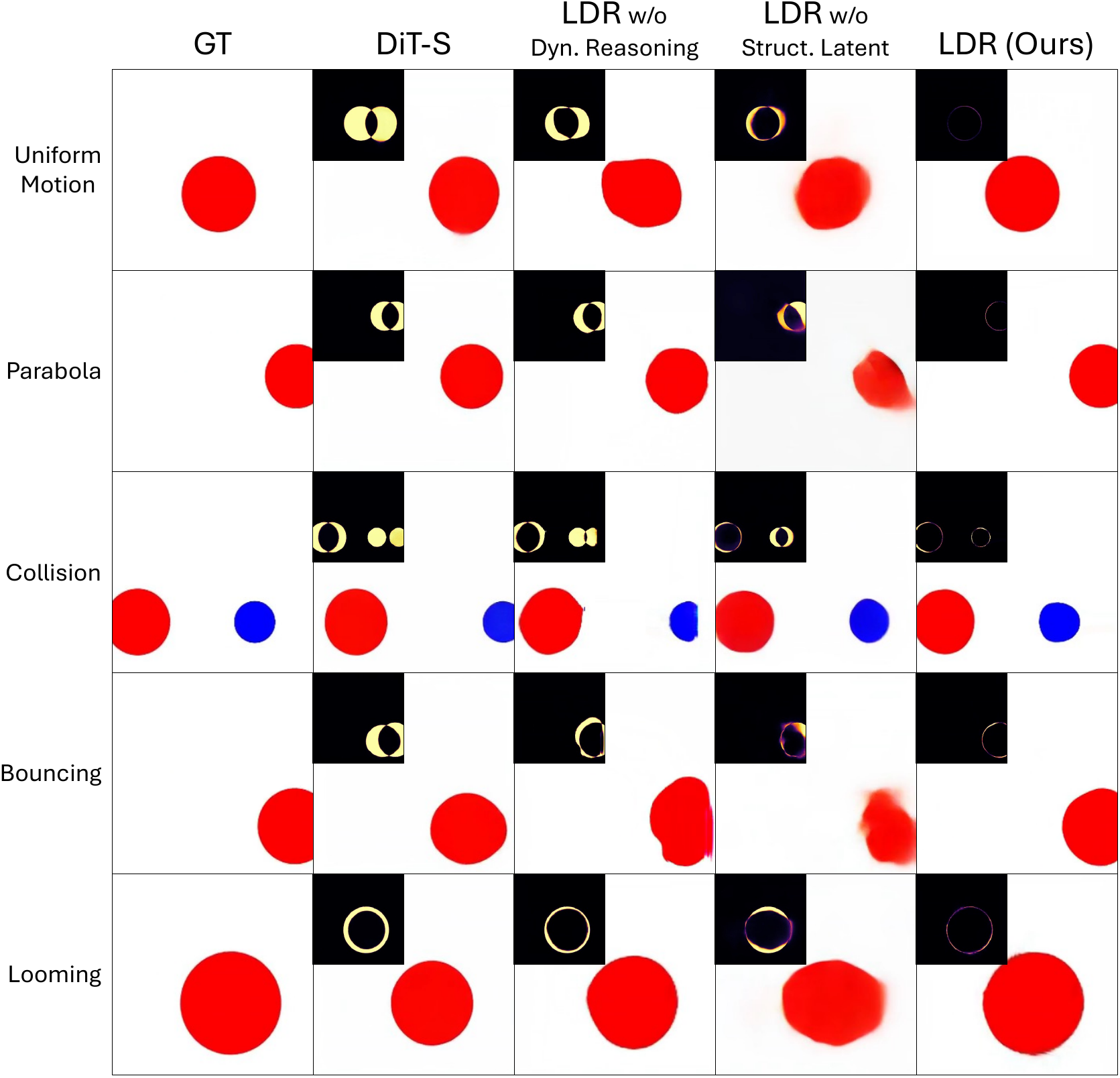}
    \vspace{-.2cm}
    \caption{
    \textbf{Qualitative OOD comparison under single-task training.}
    For each task and each method, we show a predicted frame and its error map (darker is better). 
    LDR stays close to the GT, while the baseline and both ablations drift.
    }
    \label{fig:qualitative-s}
    \vspace{-.2cm}
\end{figure}

%% file: Tabs/main_256_single.tex
\begin{table*}[t]
\centering
\scriptsize
\setlength{\tabcolsep}{7.2pt}
\renewcommand{\arraystretch}{0.8}
\begin{tabular}{cl ccc ccc ccc ccc}
\toprule
\multirow{2}{*}{Task} & \multirow{2}{*}{Metric} & \multicolumn{3}{c}{DiT-S (PhyWorld Baseline)} & \multicolumn{3}{c}{LDR w/o Dyn.\ Reasoning} & \multicolumn{3}{c}{LDR w/o Struct.\ Latent} & \multicolumn{3}{c}{\textbf{LDR (Ours)}} \\
\cmidrule(lr){3-5}\cmidrule(lr){6-8}\cmidrule(lr){9-11}\cmidrule(lr){12-14}
& & ID & OOD & Gap & ID & OOD & Gap & ID & OOD & Gap & ID & OOD & Gap \\
\midrule
Uniform & \texttt{pos} & 0.081 & 0.705 & 0.624 & \underline{0.066} & 0.297 & 0.232 & 0.082 & \underline{0.107} & \underline{0.025} & \textbf{0.044} & \textbf{0.046} & \textbf{0.003} \\
 & \mygray{\texttt{rad}} & \mygray{0.015} & \mygray{\underline{0.026}} & \mygray{\underline{0.011}} & \mygray{0.018} & \mygray{0.045} & \mygray{0.027} & \mygray{\textbf{0.011}} & \mygray{0.064} & \mygray{0.053} & \mygray{\underline{0.014}} & \mygray{\textbf{0.020}} & \mygray{\textbf{0.006}} \\
 Parabola & \texttt{pos} & 0.076 & 0.459 & 0.383 & 0.134 & \underline{0.191} & \underline{0.057} & \underline{0.054} & 0.200 & 0.146 & \textbf{0.031} & \textbf{0.040} & \textbf{0.009} \\
 & \mygray{\texttt{rad}} & \mygray{\textbf{0.009}} & \mygray{0.040} & \mygray{0.031} & \mygray{0.024} & \mygray{\textbf{0.024}} & \mygray{\textbf{0.000}} & \mygray{\underline{0.013}} & \mygray{0.084} & \mygray{0.071} & \mygray{0.019} & \mygray{\textbf{0.024}} & \mygray{\underline{0.005}} \\
 Collision & \texttt{pos}$_\text{F}$ & 0.057 & 0.191 & 0.134 & 0.126 & 0.327 & 0.201 & \textbf{0.049} & \underline{0.106} & \underline{0.057} & \underline{0.051} & \textbf{0.078} & \textbf{0.027} \\
 & \mygray{\texttt{rad}$_\text{F}$} & \mygray{0.062} & \mygray{\textbf{0.069}} & \mygray{\textbf{0.007}} & \mygray{0.066} & \mygray{0.107} & \mygray{0.041} & \mygray{\textbf{0.052}} & \mygray{0.107} & \mygray{0.055} & \mygray{\underline{0.053}} & \mygray{\underline{0.078}} & \mygray{\underline{0.025}} \\
 & \texttt{pos}$_\text{P}$ & 0.070 & 0.333 & 0.263 & 0.176 & 0.521 & 0.345 & \textbf{0.053} & \underline{0.151} & \underline{0.098} & \underline{0.063} & \textbf{0.104} & \textbf{0.041} \\
 & \mygray{\texttt{rad}$_\text{P}$} & \mygray{0.071} & \mygray{\textbf{0.079}} & \mygray{\textbf{0.008}} & \mygray{0.079} & \mygray{0.132} & \mygray{0.053} & \mygray{\textbf{0.061}} & \mygray{0.125} & \mygray{0.064} & \mygray{\textbf{0.061}} & \mygray{\underline{0.094}} & \mygray{\underline{0.033}} \\
 Bouncing & \texttt{pos} & \underline{0.064} & 0.243 & 0.179 & 0.093 & 0.439 & 0.346 & 0.070 & \underline{0.180} & \underline{0.110} & \textbf{0.057} & \textbf{0.075} & \textbf{0.018} \\
 & \mygray{\texttt{rad}} & \mygray{0.020} & \mygray{\underline{0.043}} & \mygray{\underline{0.023}} & \mygray{0.029} & \mygray{0.083} & \mygray{0.054} & \mygray{\textbf{0.013}} & \mygray{0.099} & \mygray{0.086} & \mygray{\underline{0.015}} & \mygray{\textbf{0.024}} & \mygray{\textbf{0.009}} \\
 Looming & \texttt{pos} & 0.067 & 0.246 & 0.179 & 0.104 & \underline{0.109} & \textbf{0.005} & \underline{0.053} & 0.163 & 0.110 & \textbf{0.040} & \textbf{0.046} & \underline{0.006} \\
 & \texttt{rad} & 0.029 & 0.169 & 0.140 & 0.021 & 0.062 & 0.041 & \underline{0.020} & \underline{0.047} & \underline{0.027} & \textbf{0.015} & \textbf{0.029} & \textbf{0.014} \\
\midrule
\rowcolor{gray!12}\textbf{Avg} & \texttt{pos} & 0.069 & 0.369 & 0.300\,\scriptsize(23.9$\times$) & 0.104 & 0.273 & 0.168\,\scriptsize(13.4$\times$) & \underline{0.062} & \underline{0.151} & \underline{0.090}\,\scriptsize(7.2$\times$) & \textbf{0.044} & \textbf{0.057} & \textbf{0.013} \\
\rowcolor{gray!12}\textbf{Avg} & \mygray{\texttt{rad}} & \mygray{0.027} & \mygray{0.069} & \mygray{0.042\,\scriptsize(3.6$\times$)} & \mygray{0.032} & \mygray{\underline{0.064}} & \mygray{\underline{0.032}\,\scriptsize(2.8$\times$)} & \mygray{\textbf{0.022}} & \mygray{0.080} & \mygray{0.058\,\scriptsize(4.9$\times$)} & \mygray{\underline{0.023}} & \mygray{\textbf{0.035}} & \mygray{\textbf{0.012}} \\
\bottomrule
\end{tabular}
\vspace{-.2cm}
\caption{
\textbf{Quantitative comparison under single-task training at $256^2$.}
We report position (\texttt{pos}) and radius (\texttt{rad}) errors on both ID and OOD splits, extracted from the predicted frames, and the ID-OOD gap: $\max(0,\text{OOD}{-}\text{ID})$.
For collision, we report both full-window (F) and post-collision (P) results.
The \textbf{Avg} rows average results across the five tasks (use only the full-window result for collision), and each competitor's gap is annotated with its ratio to LDR's ($\times$).
Per split, the best number is in \textbf{bold} and the second best is \underline{underlined}.
\mygray{Gray rows} mark numbers that do not reflect the motion: looming's \texttt{rad} reflects the growing or shrinking motion of the ball, but in the other tasks the ball's radius is physically constant, so \texttt{rad} there reflects primarily rendering fidelity rather than dynamics.
}
\label{tab:main:256:single}
\end{table*}

%% file: Tabs/main_256_joint.tex
\begin{table*}[t]
\centering
\scriptsize
\setlength{\tabcolsep}{7.2pt}
\renewcommand{\arraystretch}{0.8}
\begin{tabular}{cl ccc ccc ccc ccc}
\toprule
\multirow{2}{*}{Task} & \multirow{2}{*}{Metric} & \multicolumn{3}{c}{DiT-S (PhyWorld Baseline)} & \multicolumn{3}{c}{LDR w/o Dyn.\ Reasoning} & \multicolumn{3}{c}{LDR w/o Struct.\ Latent} & \multicolumn{3}{c}{\textbf{LDR (Ours)}} \\
\cmidrule(lr){3-5}\cmidrule(lr){6-8}\cmidrule(lr){9-11}\cmidrule(lr){12-14}
& & ID & OOD & Gap & ID & OOD & Gap & ID & OOD & Gap & ID & OOD & Gap \\
\midrule
Uniform & \texttt{pos} & 0.078 & 0.615 & 0.537 & 0.885 & 0.993 & 0.108 & \underline{0.056} & \underline{0.101} & \underline{0.045} & \textbf{0.044} & \textbf{0.046} & \textbf{0.002} \\
 & \mygray{\texttt{rad}} & \mygray{\textbf{0.017}} & \mygray{\underline{0.048}} & \mygray{\underline{0.031}} & \mygray{0.036} & \mygray{0.080} & \mygray{0.044} & \mygray{0.019} & \mygray{0.051} & \mygray{0.032} & \mygray{\textbf{0.017}} & \mygray{\textbf{0.023}} & \mygray{\textbf{0.006}} \\
 Parabola & \texttt{pos} & 0.075 & 0.978 & 0.903 & 0.193 & \underline{0.279} & \underline{0.086} & \underline{0.060} & 0.321 & 0.261 & \textbf{0.050} & \textbf{0.067} & \textbf{0.017} \\
 & \mygray{\texttt{rad}} & \mygray{\textbf{0.015}} & \mygray{0.038} & \mygray{0.023} & \mygray{0.027} & \mygray{\underline{0.037}} & \mygray{\underline{0.010}} & \mygray{\underline{0.016}} & \mygray{0.050} & \mygray{0.034} & \mygray{\underline{0.016}} & \mygray{\textbf{0.025}} & \mygray{\textbf{0.009}} \\
 Collision & \texttt{pos}$_\text{F}$ & 0.090 & 0.235 & 0.145 & 0.202 & 0.392 & 0.190 & \underline{0.055} & \underline{0.112} & \underline{0.057} & \textbf{0.051} & \textbf{0.073} & \textbf{0.022} \\
 & \mygray{\texttt{rad}$_\text{F}$} & \mygray{\textbf{0.052}} & \mygray{\textbf{0.067}} & \mygray{\underline{0.015}} & \mygray{0.085} & \mygray{0.174} & \mygray{0.089} & \mygray{\underline{0.055}} & \mygray{0.097} & \mygray{0.042} & \mygray{0.078} & \mygray{\underline{0.079}} & \mygray{\textbf{0.001}} \\
 & \texttt{pos}$_\text{P}$ & 0.121 & 0.431 & 0.310 & 0.277 & 0.664 & 0.387 & \underline{0.062} & \underline{0.153} & \underline{0.091} & \textbf{0.060} & \textbf{0.112} & \textbf{0.052} \\
 & \mygray{\texttt{rad}$_\text{P}$} & \mygray{\textbf{0.061}} & \mygray{\textbf{0.078}} & \mygray{\underline{0.017}} & \mygray{0.104} & \mygray{0.214} & \mygray{0.110} & \mygray{\underline{0.064}} & \mygray{0.111} & \mygray{0.047} & \mygray{0.095} & \mygray{\underline{0.096}} & \mygray{\textbf{0.001}} \\
 Bouncing & \texttt{pos} & 0.101 & 0.780 & 0.679 & 0.279 & 0.626 & 0.347 & \underline{0.074} & \underline{0.240} & \underline{0.166} & \textbf{0.063} & \textbf{0.107} & \textbf{0.044} \\
 & \mygray{\texttt{rad}} & \mygray{\textbf{0.012}} & \mygray{\textbf{0.036}} & \mygray{\underline{0.024}} & \mygray{0.024} & \mygray{0.068} & \mygray{0.044} & \mygray{\underline{0.016}} & \mygray{0.056} & \mygray{0.040} & \mygray{0.022} & \mygray{\underline{0.042}} & \mygray{\textbf{0.020}} \\
 Looming & \texttt{pos} & 0.087 & 0.351 & 0.265 & 0.909 & 0.828 & \textbf{0.000} & \underline{0.051} & \underline{0.188} & 0.137 & \textbf{0.041} & \textbf{0.048} & \underline{0.007} \\
 & \texttt{rad} & 0.027 & 0.138 & 0.111 & 0.166 & 0.306 & 0.140 & \underline{0.019} & \underline{0.054} & \underline{0.035} & \textbf{0.018} & \textbf{0.032} & \textbf{0.014} \\
\midrule
\rowcolor{gray!12}\textbf{Avg} & \texttt{pos} & 0.086 & 0.592 & 0.506\,\scriptsize(27.7$\times$) & 0.494 & 0.624 & 0.146\,\scriptsize(8.0$\times$) & \underline{0.059} & \underline{0.192} & \underline{0.133}\,\scriptsize(7.3$\times$) & \textbf{0.050} & \textbf{0.068} & \textbf{0.018} \\
\rowcolor{gray!12}\textbf{Avg} & \mygray{\texttt{rad}} & \mygray{\textbf{0.025}} & \mygray{0.065} & \mygray{0.040\,\scriptsize(4.1$\times$)} & \mygray{0.068} & \mygray{0.133} & \mygray{0.065\,\scriptsize(6.5$\times$)} & \mygray{\textbf{0.025}} & \mygray{\underline{0.062}} & \mygray{\underline{0.037}\,\scriptsize(3.7$\times$)} & \mygray{0.030} & \mygray{\textbf{0.040}} & \mygray{\textbf{0.010}} \\
\bottomrule
\end{tabular}
\vspace{-.2cm}
\caption{
\textbf{Quantitative comparison under joint five-task training at $256^2$.}
Same organization, metrics, and notations as Tab.~\ref{tab:main:256:single}.
}
\vspace{-.2cm}
\label{tab:main:256:joint}
\end{table*}

%% file: Tabs/main_128_single.tex
\begin{table*}[t]
\centering
\scriptsize
\setlength{\tabcolsep}{7.0pt}
\renewcommand{\arraystretch}{0.8}
\begin{tabular}{cl ccc ccc ccc ccc}
\toprule
\multirow{2}{*}{Task} & \multirow{2}{*}{Metric} & \multicolumn{3}{c}{DiT-S (PhyWorld Baseline)} & \multicolumn{3}{c}{LDR w/o Dyn.\ Reasoning} & \multicolumn{3}{c}{LDR w/o Struct.\ Latent} & \multicolumn{3}{c}{\textbf{LDR (Ours)}} \\
\cmidrule(lr){3-5}\cmidrule(lr){6-8}\cmidrule(lr){9-11}\cmidrule(lr){12-14}
& & ID & OOD & Gap & ID & OOD & Gap & ID & OOD & Gap & ID & OOD & Gap \\
\midrule
Uniform & \texttt{pos} & 0.148 & 0.381 & 0.233 & 0.136 & 0.414 & 0.278 & \underline{0.128} & \underline{0.137} & \underline{0.009} & \textbf{0.088} & \textbf{0.087} & \textbf{0.000} \\
 & \mygray{\texttt{rad}} & \mygray{0.018} & \mygray{\underline{0.040}} & \mygray{\underline{0.022}} & \mygray{\underline{0.013}} & \mygray{0.053} & \mygray{0.040} & \mygray{0.031} & \mygray{0.091} & \mygray{0.060} & \mygray{\textbf{0.008}} & \mygray{\textbf{0.016}} & \mygray{\textbf{0.008}} \\
 Parabola & \texttt{pos} & 0.166 & 0.388 & 0.222 & \underline{0.086} & \underline{0.203} & \underline{0.117} & 0.091 & 0.301 & 0.210 & \textbf{0.081} & \textbf{0.080} & \textbf{0.000} \\
 & \mygray{\texttt{rad}} & \mygray{\textbf{0.011}} & \mygray{\underline{0.038}} & \mygray{\underline{0.027}} & \mygray{\underline{0.013}} & \mygray{0.040} & \mygray{\underline{0.027}} & \mygray{0.018} & \mygray{0.128} & \mygray{0.110} & \mygray{0.020} & \mygray{\textbf{0.026}} & \mygray{\textbf{0.006}} \\
 Collision & \texttt{pos}$_\text{F}$ & 0.104 & 0.240 & 0.136 & \textbf{0.078} & 0.205 & 0.127 & 0.086 & \underline{0.161} & \underline{0.075} & \underline{0.085} & \textbf{0.100} & \textbf{0.015} \\
 & \mygray{\texttt{rad}$_\text{F}$} & \mygray{0.059} & \mygray{\textbf{0.081}} & \mygray{\textbf{0.022}} & \mygray{\textbf{0.050}} & \mygray{0.109} & \mygray{0.059} & \mygray{\underline{0.053}} & \mygray{0.125} & \mygray{0.072} & \mygray{0.054} & \mygray{\underline{0.084}} & \mygray{\underline{0.030}} \\
 & \texttt{pos}$_\text{P}$ & 0.119 & 0.370 & 0.251 & \textbf{0.090} & 0.312 & 0.222 & \textbf{0.090} & \underline{0.218} & \underline{0.128} & 0.092 & \textbf{0.128} & \textbf{0.036} \\
 & \mygray{\texttt{rad}$_\text{P}$} & \mygray{0.067} & \mygray{\textbf{0.090}} & \mygray{\textbf{0.023}} & \mygray{\textbf{0.059}} & \mygray{0.132} & \mygray{0.073} & \mygray{\underline{0.060}} & \mygray{0.142} & \mygray{0.082} & \mygray{0.063} & \mygray{\underline{0.101}} & \mygray{\underline{0.038}} \\
 Bouncing & \texttt{pos} & 0.114 & 0.308 & 0.194 & 0.181 & 0.471 & 0.290 & \textbf{0.098} & \underline{0.261} & \underline{0.163} & \underline{0.102} & \textbf{0.114} & \textbf{0.012} \\
 & \mygray{\texttt{rad}} & \mygray{\textbf{0.013}} & \mygray{\underline{0.051}} & \mygray{\underline{0.038}} & \mygray{\underline{0.015}} & \mygray{0.089} & \mygray{0.074} & \mygray{\underline{0.015}} & \mygray{0.143} & \mygray{0.128} & \mygray{0.024} & \mygray{\textbf{0.043}} & \mygray{\textbf{0.019}} \\
 Looming & \texttt{pos} & 0.097 & 0.249 & 0.152 & 0.148 & 0.317 & 0.169 & \underline{0.088} & \underline{0.240} & \underline{0.152} & \textbf{0.074} & \textbf{0.095} & \textbf{0.021} \\
 & \texttt{rad} & 0.035 & 0.166 & 0.131 & 0.041 & 0.148 & 0.107 & \textbf{0.023} & \underline{0.053} & \underline{0.030} & \underline{0.029} & \textbf{0.045} & \textbf{0.016} \\
\midrule
\rowcolor{gray!12}\textbf{Avg} & \texttt{pos} & 0.126 & 0.313 & 0.187\,\scriptsize(19.6$\times$) & 0.126 & 0.322 & 0.196\,\scriptsize(20.5$\times$) & \underline{0.098} & \underline{0.220} & \underline{0.122}\,\scriptsize(12.7$\times$) & \textbf{0.086} & \textbf{0.095} & \textbf{0.010} \\
\rowcolor{gray!12}\textbf{Avg} & \mygray{\texttt{rad}} & \mygray{\underline{0.027}} & \mygray{\underline{0.075}} & \mygray{\underline{0.048}\,\scriptsize(3.0$\times$)} & \mygray{\textbf{0.026}} & \mygray{0.088} & \mygray{0.062\,\scriptsize(3.9$\times$)} & \mygray{0.028} & \mygray{0.108} & \mygray{0.080\,\scriptsize(5.1$\times$)} & \mygray{\underline{0.027}} & \mygray{\textbf{0.043}} & \mygray{\textbf{0.016}} \\
\bottomrule
\end{tabular}
\vspace{-.2cm}
\caption{
\textbf{Quantitative comparison under single task training at $128^2$.}
Same organization, metrics, and notations as Tab.~\ref{tab:main:256:single}.
}
\label{tab:main:128:single}
\end{table*}

%% file: Tabs/main_128_joint.tex
\begin{table*}[t]
\centering
\scriptsize
\setlength{\tabcolsep}{7.4pt}
\renewcommand{\arraystretch}{0.8}
\begin{tabular}{cl ccc ccc ccc ccc}
\toprule
\multirow{2}{*}{Task} & \multirow{2}{*}{Metric} & \multicolumn{3}{c}{DiT-S (PhyWorld Baseline)} & \multicolumn{3}{c}{LDR w/o Dyn.\ Reasoning} & \multicolumn{3}{c}{LDR w/o Struct.\ Latent} & \multicolumn{3}{c}{\textbf{LDR (Ours)}} \\
\cmidrule(lr){3-5}\cmidrule(lr){6-8}\cmidrule(lr){9-11}\cmidrule(lr){12-14}
& & ID & OOD & Gap & ID & OOD & Gap & ID & OOD & Gap & ID & OOD & Gap \\
\midrule
Uniform & \texttt{pos} & 0.094 & \underline{0.172} & \underline{0.078} & 0.800 & 0.927 & 0.127 & \textbf{0.090} & 0.193 & 0.103 & \underline{0.091} & \textbf{0.089} & \textbf{0.000} \\
 & \mygray{\texttt{rad}} & \mygray{0.027} & \mygray{\underline{0.035}} & \mygray{\underline{0.008}} & \mygray{0.033} & \mygray{0.098} & \mygray{0.065} & \mygray{\underline{0.020}} & \mygray{0.067} & \mygray{0.047} & \mygray{\textbf{0.016}} & \mygray{\textbf{0.022}} & \mygray{\textbf{0.006}} \\
 Parabola & \texttt{pos} & \underline{0.086} & 0.358 & 0.272 & 0.169 & \underline{0.294} & \underline{0.125} & 0.091 & 0.383 & 0.292 & \textbf{0.083} & \textbf{0.125} & \textbf{0.042} \\
 & \mygray{\texttt{rad}} & \mygray{\textbf{0.014}} & \mygray{\underline{0.041}} & \mygray{0.027} & \mygray{0.036} & \mygray{0.062} & \mygray{\underline{0.026}} & \mygray{\underline{0.017}} & \mygray{0.075} & \mygray{0.058} & \mygray{\underline{0.017}} & \mygray{\textbf{0.030}} & \mygray{\textbf{0.013}} \\
 Collision & \texttt{pos}$_\text{F}$ & 0.102 & 0.179 & 0.077 & 0.151 & 0.285 & 0.134 & \underline{0.093} & \underline{0.162} & \underline{0.069} & \textbf{0.086} & \textbf{0.106} & \textbf{0.020} \\
 & \mygray{\texttt{rad}$_\text{F}$} & \mygray{\textbf{0.050}} & \mygray{\textbf{0.064}} & \mygray{\textbf{0.014}} & \mygray{0.057} & \mygray{0.125} & \mygray{0.068} & \mygray{0.059} & \mygray{0.131} & \mygray{0.072} & \mygray{\underline{0.056}} & \mygray{\underline{0.079}} & \mygray{\underline{0.023}} \\
 & \texttt{pos}$_\text{P}$ & 0.130 & 0.288 & 0.158 & 0.203 & 0.462 & 0.259 & \underline{0.101} & \underline{0.218} & \underline{0.117} & \textbf{0.094} & \textbf{0.132} & \textbf{0.038} \\
 & \mygray{\texttt{rad}$_\text{P}$} & \mygray{\textbf{0.060}} & \mygray{\textbf{0.074}} & \mygray{\textbf{0.014}} & \mygray{0.069} & \mygray{0.155} & \mygray{0.086} & \mygray{0.068} & \mygray{0.147} & \mygray{0.079} & \mygray{\underline{0.066}} & \mygray{\underline{0.094}} & \mygray{\underline{0.028}} \\
 Bouncing & \texttt{pos} & \underline{0.104} & \underline{0.226} & \underline{0.122} & 0.180 & 0.471 & 0.291 & 0.105 & 0.369 & 0.264 & \textbf{0.100} & \textbf{0.162} & \textbf{0.062} \\
 & \mygray{\texttt{rad}} & \mygray{\underline{0.014}} & \mygray{\textbf{0.030}} & \mygray{\textbf{0.016}} & \mygray{0.032} & \mygray{0.087} & \mygray{0.055} & \mygray{0.020} & \mygray{0.079} & \mygray{0.059} & \mygray{\textbf{0.011}} & \mygray{\underline{0.046}} & \mygray{\underline{0.035}} \\
 Looming & \texttt{pos} & \underline{0.086} & \underline{0.172} & 0.086 & 0.852 & 0.809 & \textbf{0.000} & \textbf{0.085} & 0.260 & 0.175 & 0.090 & \textbf{0.090} & \textbf{0.000} \\
 & \texttt{rad} & 0.035 & 0.113 & 0.078 & 0.172 & 0.309 & 0.137 & \underline{0.022} & \underline{0.058} & \underline{0.036} & \textbf{0.016} & \textbf{0.037} & \textbf{0.021} \\
\midrule
\rowcolor{gray!12}\textbf{Avg} & \texttt{pos} & 0.094 & \underline{0.222} & \underline{0.127}\,\scriptsize(5.1$\times$) & 0.430 & 0.557 & 0.135\,\scriptsize(5.5$\times$) & \underline{0.093} & 0.273 & 0.181\,\scriptsize(7.3$\times$) & \textbf{0.090} & \textbf{0.114} & \textbf{0.025} \\
\rowcolor{gray!12}\textbf{Avg} & \mygray{\texttt{rad}} & \mygray{\underline{0.028}} & \mygray{\underline{0.057}} & \mygray{\underline{0.029}\,\scriptsize(1.5$\times$)} & \mygray{0.066} & \mygray{0.136} & \mygray{0.070\,\scriptsize(3.6$\times$)} & \mygray{\underline{0.028}} & \mygray{0.082} & \mygray{0.054\,\scriptsize(2.8$\times$)} & \mygray{\textbf{0.023}} & \mygray{\textbf{0.043}} & \mygray{\textbf{0.020}} \\
\bottomrule
\end{tabular}
\vspace{-.2cm}
\caption{
\textbf{Quantitative comparison under joint five-task training at $128^2$.}
Same organization, metrics, and notations as Tab.~\ref{tab:main:256:single}.
}
\vspace{-.2cm}
\label{tab:main:128:joint}
\end{table*}

%% file: Figs/Qualitative-J.tex
\begin{figure}[!th]
    \centering
    \includegraphics[width=0.99\linewidth]{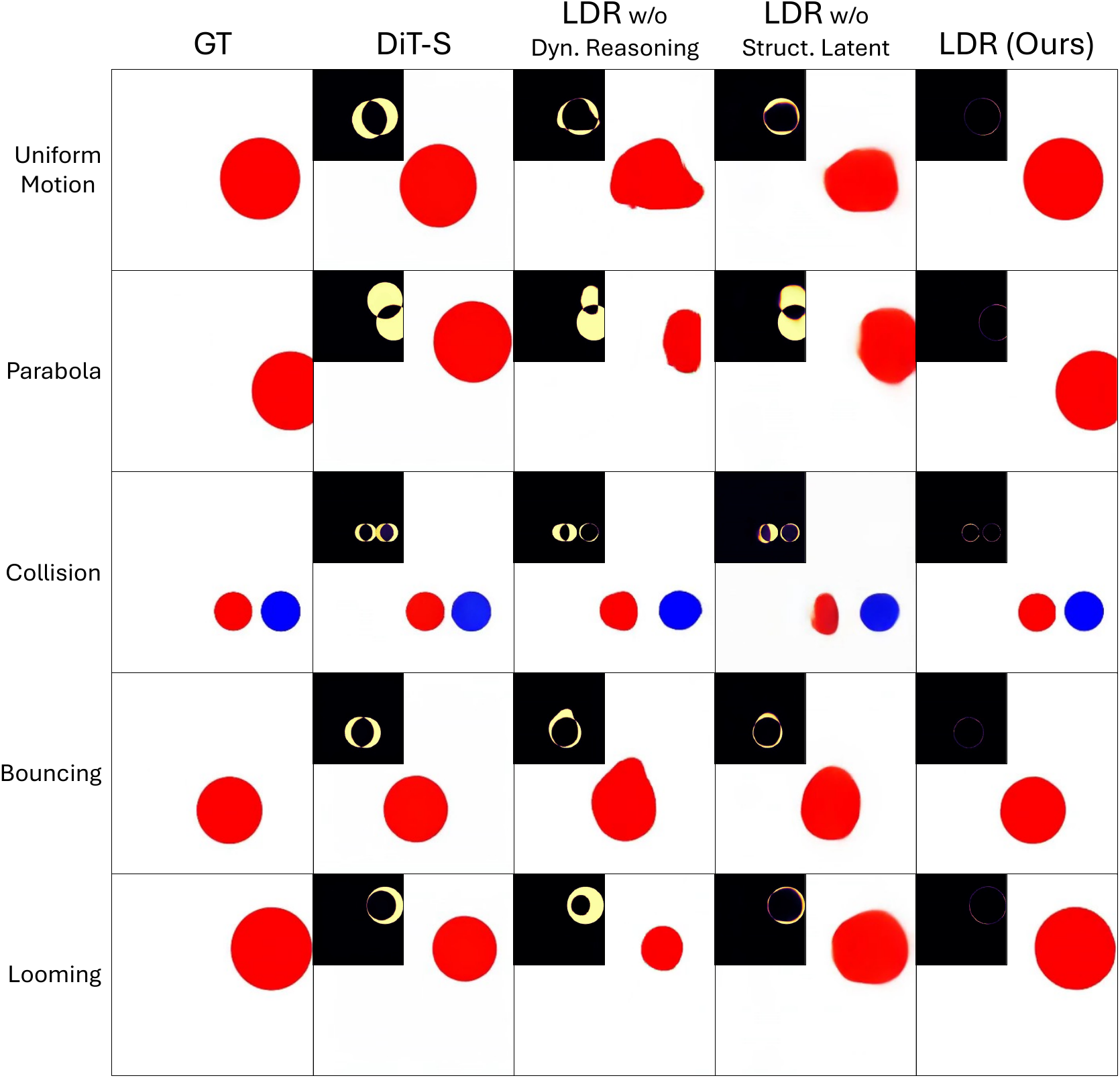}
    \vspace{-.2cm}
    \caption{
    \textbf{Qualitative OOD comparison under joint five-task training.}
    A single model holds all five tasks, yet LDR still stays close to GT while the others drift.
    }
    \label{fig:qualitative-j}
    \vspace{-.2cm}
\end{figure}

%% file: Figs/Qualitative-OSS.tex
\begin{figure}[!th]
    \centering
    \includegraphics[width=0.99\linewidth]{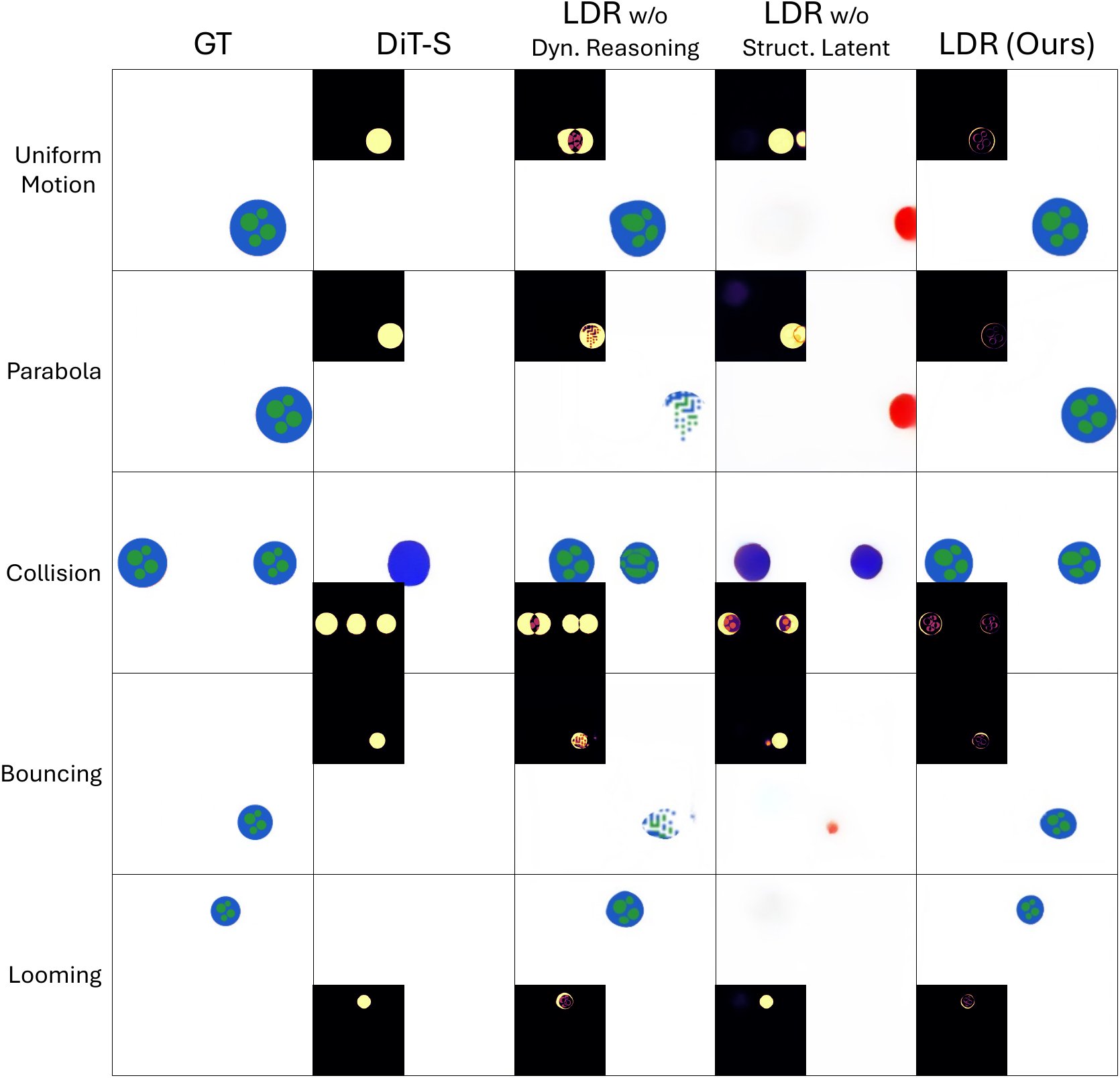}
    \vspace{-.2cm}
    \caption{
    \textbf{Stress test under severe OOD shift.}
    Trained only on red balls but tested on an unseen object (e.g., ``earth''), LDR still predicts the correct motion, while the others fail.
    }
    \label{fig:severe_shift}
    \vspace{-.2cm}
\end{figure}

%% file: Tabs/efficiency.tex
\begin{table}[t]
\centering
\scriptsize
\setlength{\tabcolsep}{11.5pt}
\renewcommand{\arraystretch}{0.8}
\begin{tabular}{lcc}
\toprule
 & DiT-S (PhyWorld Baseline) & \textbf{LDR (Ours)} \\
\midrule
Params (M)               & 106.1114 & \textbf{4.0933} \\
Ratio to LDR ($\times$)  & 25.9     & \textbf{1.0} \\
Latency @128$^2$ (s)     & 0.7451   & \textbf{0.0174} \\
Ratio to LDR ($\times$)  & 42.8     & \textbf{1.0} \\
Latency @256$^2$ (s)     & 5.2069   & \textbf{0.0363} \\
Ratio to LDR ($\times$)  & 143.4    & \textbf{1.0} \\
\bottomrule
\end{tabular}
\vspace{-.2cm}
\caption{
\textbf{Efficiency study} measured on one NVIDIA A100-80G on $32$-frame clips.
The DiT-S baseline uses $50$ DDIM steps following \cite{kang2024far}.
}
\vspace{-.2cm}
\label{tab:efficiency}
\end{table}

%% file: Secs/Summary.tex
\section{Summary}

We introduced Latent Dynamics Reasoning (LDR), an extrapolative video world model that predicts by reasoning about the latent dynamics rather than regressing future frames directly.
Reasoning about the dynamics lets a video world model learn how the world evolves and carry that knowledge beyond what it has seen.
To our knowledge, LDR is the first video world model that extrapolates learned dynamics beyond its training distribution.

\subsubsection{Limitation.}
The dynamics reasoning of LDR is content-agnostic: the kinematic integration makes no assumption about what the latent encodes or which laws of motion govern the pixels. However, the latent representation it reasons over limits the practical universality of LDR.
First, the SL encodes only the ``structure'' of the image feature rather than the ``content'' itself, so LDR may not model dynamics that live in the appearance, such as color evolving over time.
Second, the SL, implemented as a geometric coordinate, handles simple scenes but may not be expressive enough for richer scenes.
Finally, we validate LDR as a principle on simulated scenes with simple objects, and scaling it to richer, even real-world scenes with larger models remains future work.
Strengthening the latent representation while keeping the general reasoning core is a promising path toward true video world models that learn how the world evolves.